\pdfoutput=1
\documentclass[11pt]{article}

\usepackage{EMNLP2023}

\usepackage{times}
\usepackage{latexsym}

\usepackage[T1]{fontenc}
\usepackage[utf8]{inputenc}

\usepackage{microtype}

\usepackage{inconsolata}

\usepackage{xspace}
\usepackage{booktabs}

\usepackage{graphicx}
\usepackage{enumitem}
\usepackage{listings}
\definecolor{codegreen}{rgb}{0,0.6,0}
\definecolor{codegray}{rgb}{0.5,0.5,0.5}
\definecolor{codepurple}{rgb}{0.58,0,0.82}
\definecolor{backcolour}{rgb}{0.95,0.95,0.92}

\lstdefinestyle{mystyle}{
    commentstyle=\color{codegreen},
    keywordstyle=\color{magenta},
    numberstyle=\tiny\color{codegray},
    stringstyle=\color{codepurple},
    breakatwhitespace=false,         
    breaklines=true,                 
    captionpos=b,                    
    keepspaces=true,                 
    numbersep=5pt,                  
    showspaces=false,                
    showstringspaces=false,
    showtabs=false,                  
    tabsize=2
}

\usepackage{balance}

\newcommand{\sotas}{\textsc{sotastream}\xspace}
\newcommand{\infinibatch}{Infinibatch\xspace}

\newcommand{\unix}{\textsc{Unix}\xspace}
\renewcommand{\line}{\texttt{Line}\xspace}
\newcommand{\pipeline}{\texttt{Pipeline}\xspace}
\newcommand{\datasource}{\texttt{DataSource}\xspace}
\newcommand{\mixer}{\texttt{Mixer}\xspace}

\usepackage{microtype}

\title{Why are you still preprocessing your machine translation training data?}
\title{\sotas: A Streaming Approach to Machine Translation Training}

\author{\begin{tabular}{ccc} 
    Matt Post & Thamme Gowda & Roman Grundkiewicz \\
    {\bf Huda Khayrallah} & {\bf Rohit Jain} & {\bf Marcin Junczys-Dowmunt}
    \end{tabular} \\
    \\
  Microsoft}

\begin{document}
\maketitle

\begin{abstract}
Many machine translation toolkits make use of a data preparation step wherein raw data is transformed into a tensor format that can be used directly by the trainer.
This preparation step is increasingly at odds with modern research and development practices because this process produces a static, unchangeable version of the training data, making common training-time needs difficult (e.g., subword sampling), time-consuming (preprocessing with large data can take days), expensive (e.g., disk space), and cumbersome (managing experiment combinatorics).
We propose an alternative approach that separates the \emph{generation} of data from the \emph{consumption} of that data.
In this approach, there is no separate pre-processing step; data generation produces an infinite stream of permutations of the raw training data, which the trainer tensorizes and batches as it is consumed.
Additionally, this data stream can be manipulated by a set of user-definable operators that provide on-the-fly modifications, such as data normalization, augmentation or filtering.
We release an open-source toolkit, \sotas, that implements this approach:~\url{https://github.com/marian-nmt/sotastream}.  
We show that it cuts training time, adds flexibility, reduces experiment management complexity, and reduces disk space, all without affecting the accuracy of the trained models.
\end{abstract}

\section{Introduction}

\begin{figure}[t]
    \centering
    \includegraphics[width=\columnwidth]{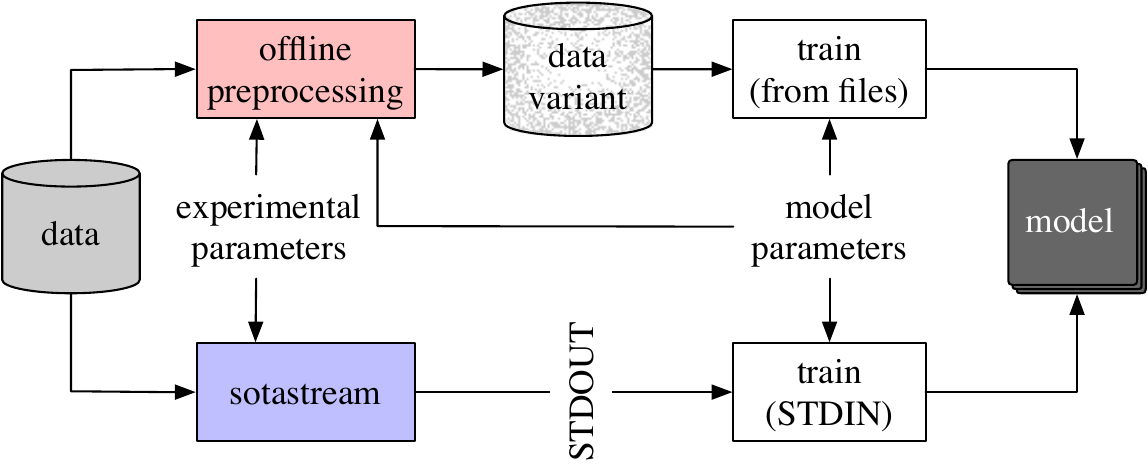}
    \caption{The \sotas approach separates data generation from consumption.
    Whereas offline tensorization requires model-specific parameters such as the vocabulary, which ties processed data to a particular training, \sotas produces data on the fly, avoiding time-consuming production and space-wasting storage of preprocessed data.}
    \label{figure:augmentibatch}
\end{figure}

A cumbersome component of training machine translation systems is working with large amounts of data.
Modern high-resource parallel datasets are often on the order of hundreds of millions of parallel sentences, and backtranslation easily doubles that \citep[Appendix A]{kocmi-etal-2022-findings}.
Because this data is too large to fit into main memory, toolkits such as \textsc{Fairseq} \cite{ott-etal-2019-fairseq} and \textsc{Sockeye} \cite{hieber-etal-2022-sockeye} make use of a preprocessing step, which transforms the training data from its raw state into a static sequence of tensors. %
These tensors can then be read in via an index and memory-mapped shards, allowing for quick assembly into batches at training time.

While this offline preprocessing prevents data loading from becoming a bottleneck in training, it creates a number of other problems:
\begin{itemize}
    \item \emph{it breaks an abstraction}: the tensorized data is tied to specific modeling decisions, such as the vocabulary;
    \item \emph{it is cumbersome}: the tensorized data cannot be changed, and even minor variations of the data must be processed separately and then managed;
    \item \emph{it is time-consuming}: pre-processing can take considerable time and must be completed before training can start; and
    \item \emph{it is wasteful}: each data variant replicates the original's disk space.
\end{itemize}
These problems exist for construction of any model, but are exacerbated in research settings, which often explore variations of the training data.

We describe an alternative that factors \emph{generation} of data from the \emph{consumption} of that data by the training toolkit.
This view presents the training data as an (infinite) stream of permutations of the raw training samples.
This stream is then consumed by the training toolkit, which tensorizes it on the fly, consuming data into a buffer from which it can assemble batches.
This framework eliminates all the problems above: variants of the data are independent of any model; arbitrary manipulations can be applied on the fly; preprocessing time is amortized over training, which can start as soon as the first batch can be constructed;and no extra disk space or management is required.

We release an open-source implementation of the proposed data generation framework called \sotas\footnote{\url{https://github.com/marian-nmt/sotastream}}.
\sotas is written in Python and uses \infinibatch\footnote{\url{https://github.com/microsoft/infinibatch}} to provide a stream of data over permutations of data sources.
It additionally provides an easily-extendable set of \emph{mixers}, \emph{augmentors}, and \emph{filters} that allow data to be probabilistically manipulated on the fly.
A particular configuration of manipulators is provided by the user in the form of a dynamically-loadable \emph{pipeline}, which defines a parameterizable recipe that can be used for training.
\sotas uses multiprocessing to reach high throughput levels that prevent starvation of the training toolkit.
And finally, it employs a standard \unix API, writing data to STDOUT.

After presenting this framework (\S~\ref{section:streams}), we conduct a quality comparison to demonstrate that it does not reduce model quality (\S~\ref{section:quality}).
We then investigate stream bandwidth under various pipelines as well as necessary toolkit consumption needs (\S~\ref{section:speed}).
We conclude by demonstrating a number of use cases (\S~\ref{section:uses}).
\section{Training from data streams}
\label{section:streams}

\begin{figure*}[t]
\begin{lstlisting}[language=Python,basicstyle=\ttfamily\small]
@pipeline("robust-case")
class RobustCasePipeline(Pipeline):
  def __init__(self, pa_dir: str, bt_dir: str, **kwargs):
    super().__init__(**kwargs)
    pa_stream = self.create_data_stream(pa_dir, processor=Augment)
    bt_stream = self.create_data_stream(bt_dir, 
      processor=partial(Augment, tag="[BT]"))  # tag the BT data
    self.stream = Mixer([pa_stream, bt_stream], self.mix_weights)
  # definitions of other class methods go here ...

def LowerCase(stream: Generator[Line]) -> Generator[Line]:
  for line in stream:
    line[0] = line[0].lower()  # lowercase the source side
    yield line

def TitleCase(stream: Generator[Line]) -> Generator[Line]:
  for line in stream:
    line[0], line[1] = line[0].title(), line[1].title()  # titlecase both sides
    yield line

def TagData(stream: Generator[Line], tag: str) -> Generator[Line]}:
  for line in stream:
    line[0] = f"{tag} {line}"  # add a target language tag to the source
    yield line

def Augment(path: str, tag: str = None) -> Generator[Line]:
  stream = UTF8File(path)   # open the path to the shard

  stream = Mixer(  # randomly mix casing variants
    [ stream, LowerCase(stream), TitleCase(stream) ],
    [ 0.95,   0.04,              0.01 ],
  )

  if tag is not None:
    stream = TagData(stream, tag)
  return stream
\end{lstlisting}
\caption{A simplified pipeline.
Streams are built by composing generator functions over input data sources (here, parallel and backtranslated data).
This example tags the backtranslated stream, then mixes it with the parallel stream using weights provided on the command line (defaulting to 1:1).
It then applies random source-lowercasing (4\%) and title-casing (1\%).}
\label{table:code}
\end{figure*}

The core idea underlying \sotas is to cleanly separate data generation from consumption of that data during training.
The \emph{data generator} is responsible for producing training samples, and the \emph{trainer} consumes them.
This factorization allows us to separate properties of the data (such as their sources, mixing ratios, and augmentations) from properties of training and the model (such as tensor format, batch size, and so on).

The current approach relies on standard \unix I/O pipes as an interface between these two pieces.
However, \sotas could also be used to generate data for offline uses, or modified to allow consumption through some other API, such as a library call that returns a generator.

\subsection{Data generation}
\label{section:generation}

\sotas is a data generator.
At a high level, it works by defining a \emph{pipeline}.
This pipeline reads from a set of zero or more input data sources, applies any augmentations, and produces a single output stream.

\paragraph{Pipelines} Pipelines are implemented by inheriting from the base \pipeline class.
The class implementation is responsible for defining the input data sources, reading from them, applying augmentations, and returning a single output stream.
These are depicted in Figure~\ref{table:code}, a simplified presentation that elides other support features, such as providing the mixing weights for the input data sources.

The pipeline has three basic components:
\begin{enumerate}
    \item Build a stream for each input data source;
    \item apply a sequence of augmentors; and
    \item merge the streams to a single output stream.
\end{enumerate}

\paragraph{Data sources}

\sotas uses \infinibatch to return a generator over a permutation of the samples in a data source.
Each \datasource object receives two key arguments: a file path to the data source on disk, $d$, and a processor function, $f$, to read it.
This can be seen in Figure~\ref{table:code} in the call to \verb|create_data_stream|$(d,f)$.

The data is received as a path to a directory of compressed TSV file shards.
\infinibatch requires that data be presented in this way.\footnote{\sotas can also receive a path to a single compressed TSV file, in which case it splits the file into shards under a temporary directory.
The default shard size is 1m lines.
Note that in this setting, it is duplicating the data, but it is a duplication of the \emph{raw} data, prior to any transformations.
The results of this automatic sharding are cached using an MD5 checksum.}
For each data epoch, \infinibatch produces a permutation of these shards.
The shards are then passed, in turn, to the function $f$, which is responsible for opening, reading, and processing the shard.
It is important to note that \infinibatch provides an \emph{infinite stream} of data; that is, it will present an infinite stream over its input data, subject to the constraint that no shard (within a data source) will be seen $n+1$ times until all shards have been seen $n$ times.
(See the Multiprocessing paragraph below for important caveats related to multiprocessing and MPI training).

\paragraph{Augmentations}

\begin{figure}
    \centering
    \begin{lstlisting}[language=python,basicstyle=\ttfamily\small]
class Line:
  def __init__(self, line: str):
    if line is not None:
      self.fields = line.split("\t")
    else:
      self.fields = []
\end{lstlisting}
    \caption{The (simplified) \line object, a lightweight wrapper around a single row of tab-separated input data.}
    \label{figure:line}
\end{figure}

The second argument to \verb|create_data_stream| is a generator function, $f$, an \infinibatch primitive whose task is to open each shard and produce an output data stream.
The output is in the form of \line objects (Figure~\ref{figure:line}, each of which is a class representation of the TSV input.
By convention in machine translation, fields 0 and 1 are treated as \emph{source} and \emph{target} segments, respectively, but the code itself makes no such assumptions.

The function is not limited in just reading and returning the data.
A key feature of \sotas is augmentations, which are arbitrary manipulations of a data stream that are easy to stack and accumulate.
This is accomplished by composing generators.
Figure~\ref{table:code} contains a number of examples in the \texttt{Augment} function.
It first opens a stream on a path (passed from \infinibatch, containing a path to a sharded file name).
It then applies lowercasing and title-casing to the input stream probabilistically, using a \mixer class to select among them with specified weights.
Finally, it prepends a tag to the data, if requested by the caller.

\paragraph{Outputting the stream}
Finally, at the top level, the (augmented) streams from different data sources are merged into a single stream.
This works in the same way as the above \mixer class example.
One additional feature is that the \pipeline class provides the ability to set these top-level data weights from the command line (\verb|--mix-weights|).

\subsection{Data consumption}

The main requirements for the trainer are to consume data into a pool, apply subword processing, organize into batches, and run backpropagation against the training objective.
Because these are done on the fly, rather than in preprocessing, special considerations must be implemented to ensure that this extra processing does not become a bottleneck for training.

In Section~\ref{section:speed}, we experiment with an implementation in the Marian toolkit \cite{junczys-dowmunt-etal-2018-marian-cost}.
Marian makes use of multiple worker threads, which pre-fetch data from STDIN into an internal memory pool, where the data is tokenized and integerized.
When the pool is filled, it is sorted and batched (according to run-time settings).
In the meantime, prefetching continues into a second pool.
As training proceeds, these two pools are used alternately for filling via prefetching and batch generation.

\subsection{Multiprocessing}

In order to sustain a sufficient throughput, \sotas makes use of multiprocessing.
This can be increasingly important if the augmentations applied are expensive to compute.
We quantify the effects of multiprocessing for generation under a handful of pipelines of varying complexity in Section~\ref{section:speed}.

Internally, this is accomplished with the \verb|multiprocessing| library.
We create separate subprocesses, each of which is provided with independent access to the data sources.
The parent process maintains a pipe to each subprocess, and queries them in sequence, reading a fixed number of lines from each in turn, and passing them to the standard output.

An important issue is raised when working with subprocesses.
If each subprocess were to return an independent permutation over the input data, merging subprocesses would not itself result in a permutation.
To address this, each of $n$ subprocesses is initalized with $\frac{1}{n}$ of the data shards, themselves assigned in round-robin order across the subprocesses.
In this way, we guarantee a permutation in settings where the number of processes evenly divides the number of shards.

When working over MPI, no such coordination takes place.
Each MPI instantiation will receive a different randomly-seeded shard permutation.

\section{Experimental setup}
\label{section:experiment}

Our experimental goal is to demonstrate that the many advantages of \sotas do not come at a cost in accuracy (\S~\ref{section:quality}) or speed (\S~\ref{section:speed}).
We do this by comparing to a number of other data loading methods.
In order to isolate the effects of changing the data loader, we conduct all of our experiments within the Marian toolkit.
Marian does not support offline data preprocessing; instead, we compare a number of different streaming settings that cover best-case scenarios.

\subsection{Streaming variations}

We compare the following data-loading variations.
\begin{itemize}
    \item \emph{Full loading}. 
    In this scenario, the trainer has direct memory access to the entire data source.
    For our experiments, Marian loads the complete datasets into main memory.
    There is some startup cost, after which all access to the data is immediate.
    \item \emph{Sequential streaming}. 
    In this approach, the training data is read sequentially, in a loop over the entire training set.
    Data is prefetched into a pool of a specified size, from which mini-batches are assembled.
    Since data is read sequentially, there is no randomization across data epochs. %
    The pool size determines an upper bound on memory usage.
    \item\emph{Randomized sequential streaming}.
    In this variant of sequential streaming, the lines in each data source are randomly permuted prior to being read.
    \item\emph{\sotas}.
    Our \infinibatch-based streaming approach.
\end{itemize}
For toolkits that support preprocessing, it is typical to construct an index, which organizes the pre-sorted and tensorized data into memory-mappable shards.
Marian does not have a preprocessing option, which means that we have no comparison to a setting where tensorization is done offline.
We thus consider full-loading to be the closest equivalent, since preprocessing is in fact a stand-in for full loading.
This can only possible affect speed comparisons (\S~\ref{section:speed}).

\subsection{Model parameters}

We conduct experiments in a large-data and small-data setting.
Our large-data setting is English--German.
We train on 297m lines of Paracrawl v9 \cite{banon-etal-2020-paracrawl} from WMT22 \cite{kocmi-etal-2022-findings}.
We use a 32k shared unigram subword model \cite{kudo-2018-subword} using SentencePiece \cite{kudo-richardson-2018-sentencepiece}, trained jointly over both sides.
We train a standard base Transformer model \cite{vaswani-etal-2017-attention} with 6/6 encoder/decoder layers, an embedding size of 1024, a feed-forward size of 4096, and 8 attention heads.
The large model is trained for 20 virtual epochs.
Since there are roughly 7.4 billion target-side tokens after tokenizing the data, this equates to roughly three passes over the data.

For the small-data setting, we train on Czech--Ukrainian, also from WMT22.
This dataset has roughly 12m parallel lines.
We use the same model and parameter settings, but train for only five virtual epochs, or roughly 30 data epochs, since the model converges by then.

\subsection{Evaluation}

Evaluation is conducted on the WMT21/en-de and WMT22/cs-uk test sets.
We use a number of metrics to capture variation:
\begin{itemize}
    \item \textbf{BLEU} \cite{papineni-etal-2002-bleu} and \textbf{chrF} \cite{popovic-2015-chrf}, both computed with sacrebleu\footnote{Version 2.3.1 with default settings.} \cite{post-2018-call}.
    \item \textbf{COMET20/22} \cite{rei-etal-2020-comet}, using model \verb|wmt20-comet-da| (EN-DE) or \verb|wmt22-comet-da| (CS-UK).

\end{itemize}

\section{Quality Comparison}
\label{section:quality}

\begin{table*}[ht]
    \small
    \centering
    \begin{tabular}{l|rrr|rrr}
    \toprule
    & \multicolumn{3}{c}{English--German (newstest2021)}
    & \multicolumn{3}{c}{Czech--Ukranian (wmttest2022)}
    \\
    Model 
    & COMET20 & BLEU & chrF
    & COMET22 & BLEU & chrF \\    
    \midrule
    Best constrained %
    & $54.8 \phantom{\pm0.0}$
    & $31.3 \phantom{\pm0.0}$
    & $60.7 \phantom{\pm0.0}$

    & $91.6 \phantom{\pm0.0}$
    & $34.7 \phantom{\pm0.0}$
    & $61.5 \phantom{\pm0.0}$
    \\
    \midrule
    Full loading  
    & $55.9 \pm0.4$
    & $34.9 \pm0.1$
    & $62.0 \pm0.0$

    & $85.5 \pm0.2$
    & $27.9 \pm0.4$
    & $55.6 \pm0.2$    
    \\
    Sequential streaming
    & $56.1 \pm0.2$
    & $35.0 \pm0.2$
    & $62.1 \pm0.0$

    & $86.4 \pm0.1$
    & $28.7 \pm0.3$
    & $56.6 \pm0.2$
    \\
    Randomized streaming
    & $55.8 \pm0.2$
    & $35.1 \pm0.0$
    & $62.2 \pm0.0$

    & $85.6 \pm0.1$
    & $27.8 \pm0.0$
    & $55.6 \pm0.2$
    \\

    \sotas
    & $55.9 \pm0.1$
    & $34.9 \pm0.1$
    & $62.1 \pm0.1$

    & $85.7 \pm0.2$
    & $28.5 \pm0.4$
    & $56.2 \pm0.2$
    \\
    \bottomrule
    \end{tabular}
    \caption{Mean over three runs for our high- and low-resource scenarios.
    The best constrained system is WeChat-AI \cite{zeng-etal-2021-wechat} for EN-DE and AMU \cite{nowakowski-etal-2022-adam} for CS-UK.}
    \label{table:ende}
\end{table*}

Table~\ref{table:ende} contains metric results for both our high- and low-resource settings.
For English--German, we observe rough equivalence across all training methods and metrics, which establishes \sotas as a viable data preparation tool.
A similar pattern holds for Czech--Ukrainian, except for the odd outlier of the sequential streaming approach.
This approach simply `cat`ed the training data repeatedly until model convergence, roughly five virtual epochs or 30 data epochs.
This result is strongest for COMET and less pronounced for BLEU and chrF.
We have no clear explanation for this; one guess is that in smaller data settings, with no filtering, curriculum effects may be more pronounced, and this is the only data generation approach with no randomization.
Among approaches that permuted the data, \sotas is on par with the others.

\section{Speed}
\label{section:speed}

Next we ask whether \sotas has a negative effect on speed.
The short answer is that it does not.
We examine speed in three settings: generation speed (\S~\ref{sec:throughput}), Marian's consumption speed (\S~\ref{sec:consumption}), and total runtime (\S~\ref{sec:total-run-time}).

\subsection{Data generation}
\label{sec:throughput}

We first examine how fast \sotas can write data to STDOUT.

Our benchmark consists of a producer and a consumer connected by UNIX pipe.
The producer varies among the tools we compare in our benchmark (described below), while the consumer is a lightweight script, whose sole purpose is to count records from STDIN and report the yield rate (the number of lines per second).
All benchmarks are run one at a time, on the same machine,\footnote{An Intel Xeon E5-2620 CPU with 32 cores, 660 GB of RAM, and running Ubuntu 20.04 LTS.} with no other CPU- or I/O-intensive processes are competing for resources.
We run each benchmark multiple times and report the average.

\begin{figure}[t]
    \centering
    \includegraphics[width=0.98\columnwidth,trim={1.85cm 6.4cm 1.85cm 7.4cm},clip]{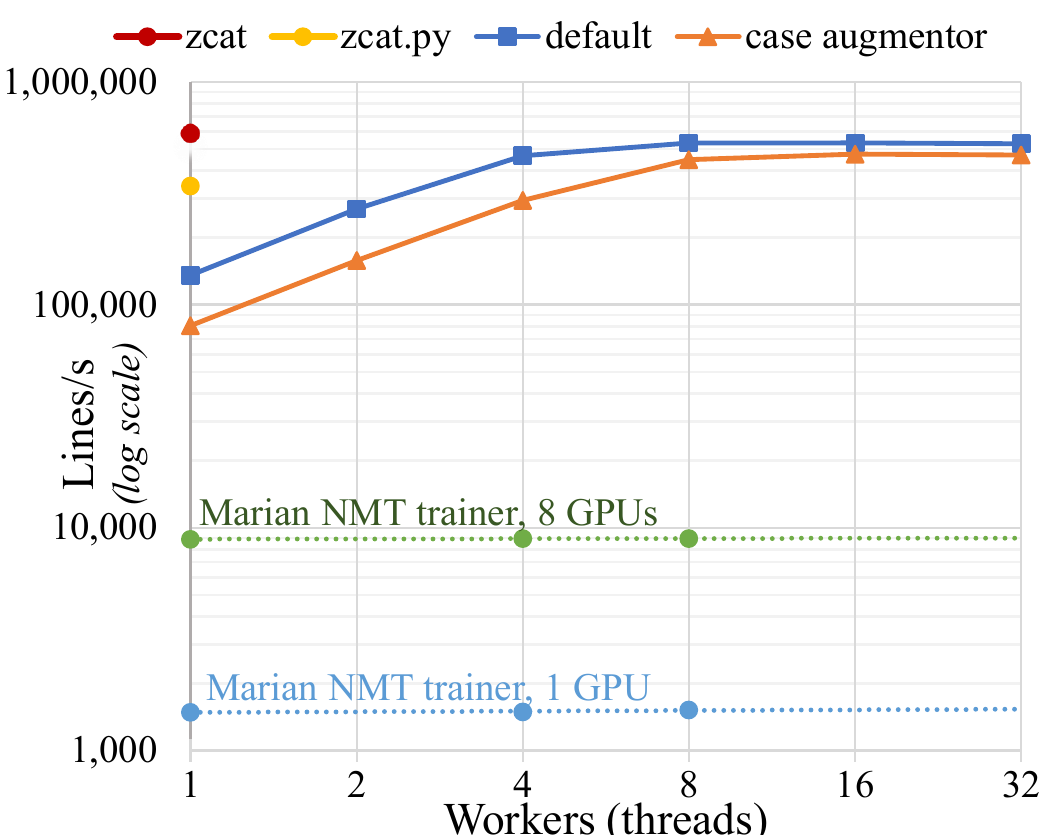}
    \caption{Generation and consumption rates with \sotas and Marian, respectively.
    \sotas is not a bottleneck, but is easily able to generate data and transmit it through the POSIX API to sustain training.}
    \label{fig:io-throughput}
\end{figure}

We compare the following generation tools:
\begin{itemize} %
    \item \texttt{zcat}: A wrapper to GNU gzip\footnote{\url{https://git.savannah.gnu.org/cgit/gzip.git/tree/zcat.in}} that decompresses and outputs lines. 
    This serves as the best case scenario, where the producer is implemented in an efficient way (e.g. C/C++) and has no time-consuming augmentations.
    
    \item \texttt{zcat.py}: Similar to \texttt{zcat}, but based on \texttt{gzip} API from Python's standard library.\footnote{\url{https://docs.python.org/3/library/gzip.html}}

    \item \texttt{default} pipeline: \sotas' default, returning lines from a single data source (\S~\ref{section:generation}) with no augmentations.

    \item \texttt{case augmentor} pipeline: the pipeline from Figure~\ref{table:code}.
    It mixes two data sources, applies case transformations, and prepends a "[BT]" tag to the backtranslated data.

\end{itemize}
We benchmark multiple worker subprocesses: $n\in\{1,2,4,8,16,32\}$.
The throughput measured is as lines/s and is given Figure~\ref{fig:io-throughput}. 
\texttt{zcat}, being the fastest, yields over 585k lines/s, and Python's alternative (\texttt{zcat.py}) yields 342k lines/s.\footnote{Measured on CPython v3.11; prior versions of CPython are found to be slower.} 
Our \sotas \texttt{default} with a single worker yields approximately 136k lines/s, which can be increased with more workers and plateaus after a certain rate (possibly due to a bottleneck in number of parallel reads supported by underlying storage device).
As we add more augmentations and mixture processes, we observe a lower yield rate than no-augmentation baselines (expected). 
However, yield rate can be improved with more worker processes. 

\subsection{Consumption}
\label{sec:consumption}

We have shown the rate at which \sotas can generate data. 
In this section, we show the rate at which one particular NMT trainer (Marian) consumes training data.
Training time and the consumption rate varies based the size of model being trained, and the number of GPUs used for training.

We train smaller Transformer models than used for Table~\ref{table:ende}, since smaller models train faster and therefore have higher data consumption needs.
We use 6/6 encoder/decoder layers, 512-dimensional embeddings, and feedforward sublayers of size 2048.
We report consumption rate for six settings: one vs.\ eight GPUs,\footnote{NVIDIA Tesla V100s with 32GB.} and using one, four, or eight prefetching worker threads.
As shown in Figure~\ref{fig:io-throughput}, a trainer with single GPU consumes about 1523 lines/s, and with eight GPUs, the consumption rate increases to 8957 lines/s. 
Even in the best case scenario (smaller model, more GPUs, and more prefetcher threads), the consumption rates of training process are lower than \sotas production rate.

We recommend running multiple workers when augmentations are slow in order to maintain sufficient output rates.
We do not experiment with them here, but in multi-node training settings coordinated with MPI, one (multiprocess) instance of \sotas should be run per node.

\subsection{Total time to run}
\label{sec:total-run-time}

Table~\ref{table:training-time} verifies that \sotas's amortized approach is neither slower nor faster than other approaches when total runtime is considered.

\begin{table}[t]
    \centering
    \small
    \begin{tabular}{l|r} 
    \toprule
    Model               &   Time (Hours) \\
    \toprule

Full loading & 36.84 $\pm$0.16 \\
Sequential streaming       & 35.51 $\pm$0.15 \\
Randomized streaming   & 35.73 $\pm$0.05 \\ 
\sotas & 35.86 $\pm$0.27 \\
\bottomrule

 \hline
    \end{tabular}
    \caption{End-to-end training time.} %
    \label{table:training-time}
\end{table}

\section{Example Use Cases}
\label{section:uses}

In this section we show example use cases how \sotas can be used to simply and easily modify data on the fly.
This provides all the advantages of training for robustness without the cumbersome task of generating (and managing) data that has been preprocessed in many different forms, which are combinatorial and impose high costs on the complexity of managing training runs.

\subsection{Mixing multiple streams of data}

Training machine translation models often requires combining different data sets in desired proportions in order to balance their size or quality or other properties.
The example in Figure~\ref{table:code} demonstrates that combining original parallel data and back-translated data can be efficiently achieved in \sotas by mixing multiple data streams with specific data weighting.
The weights for each data stream can be then specified using the command-line options: 

\begin{lstlisting}[basicstyle=\ttfamily\small]
sotastream robust-case \
  parallel.tsv.gz backtrans.tsv.gz \
  --mix-weights 1 1
\end{lstlisting}
The weights are normalized and used as probabilities with the \verb|Mixer| augmentor.

This approach, when compared to the traditional offline preparation of the data, is much simpler, more scalable, saves disk space and does not require complicated ratio-computation and data over- or downsampling.

\subsection{Data augmentation for robustness} 

\sotas{}'s augmentors provide a flexible framework for developing different methods for data augmentation, for example, case manipulation for robustness against different casing variants of the input text.
It is demonstrated in the example in Figure~\ref{table:code}, where \texttt{LowerCase} is an augmentor that lowercases the source text, and \texttt{TitleCase} converts both source and target sides to the English title-cased format.
The frequency of each variant is easily controlled with the same \verb|Mixer| used to join separate data sources.
The on-the-fly approach simplifies experiments when testing multiple variations, which is often needed in order to find optimal augmentation methods and ratios, it minimizes the burden of experiment management.

Many other types of robustness augmentation \cite{li-etal-2019-findings}, such as source-side punctuation removal, spelling errors generation, etc., can be implemented in a similar way.

\subsection{Filtering bad data examples}

In \sotas{} it is straightforward to do data filtering on the fly.
This type of filtering is especially useful in scenarios in which external data is used for model training or fine-tuning that cannot be manually filtered in a controlled way.

For example, a \texttt{URLFilter} filter that removes lines that have unmatched URLs between the source and target fields can be implemented using the provided \texttt{MatchFilter}:

\begin{lstlisting}[language=python,basicstyle=\ttfamily\small]
def URLFilter(stream):
  pattern = r'\bhttps?:\S+[a-z]\b'
  return MatchFilter(stream, pattern)
\end{lstlisting}

\subsection{Subword tokenization sampling}

The boundary separating data generation from consumption can be blurred.
For example, instead of producing raw text output, the tool could generate subwords, if provided with a subword model.
This facilitates randomized sampling of different subword segmentations from a Unigram LM model with SentencePiece's Python wrapper:

\begin{lstlisting}[language=python,basicstyle=\ttfamily\small]
import sentencepiece as sp
spm = sp.SentencePieceProcessor(
      model_file=SPM_VOCAB)

def spm_enc(stream, spm, fields=[0, 1]):
  for line in stream:
    for field in fields:
      line[field] = spm.encode(
          line[field], out_type=str,
          enable_sampling=True))
    yield line
\end{lstlisting}

\subsection{Training document-context models}

When training document models (e.g., \citet{post-junczys-dowmunt-2023-escaping}), we can easily construct pseudo-documents on the fly if the training data is augmented with a document identifier field:
\begin{lstlisting}[language=python,basicstyle=\ttfamily\small]
def read_docs(stream):
    doc, previd = [], None
    for line in stream:
        docid = line[2]
        if len(doc) and docid != previd:
            yield doc
            doc = []
        doc.append(line)
        previd = docid
    if len(doc):
        yield doc
\end{lstlisting}

\noindent A wrapper around this function could merge the source and target sides of the \verb|Line| object, perhaps subject to parameters such as a maximum sequence length, a maximum number of sentences, and structural tokens to be used as affixes.

\subsection{Alignments and other data types}

\sotas has been primarily designed for machine translation, which requires providing source and target texts as separate fields.
Other data types or metadata can be generated on the fly or provided as additional fields in the input stream.
By design the existing augmentors pass forward the unused fields, which makes introducing new fields that are used only by a subset of augmentors simple.

The example below demonstrates on-the-fly generation of word alignment using SimAlign \cite{jalili-sabet-etal-2020-simalign}:

\begin{lstlisting}[language=python,basicstyle=\ttfamily\small]
import simalign as sa
aln = sa.SentenceAligner()

def align(stream, aln, fields=[0, 1]):
  i, j = fields
  for line in stream,
    res = aln.get_word_aligns(line[i],
                              line[j])
    res = " ".join(f"{p[0]}-{p[1]}"
                   for p in res['mwmf'])
    line.append(res)
    yield line
\end{lstlisting}

\noindent The word alignment can be used directly by the trainer, e.g., for guided alignment training \cite{chen-etal-2016-guided}, or used by subsequent augmentors that may require it, e.g., constrained terminology translation annotations \cite{bergmanis-pinnis-2021-facilitating}.

\subsection{Integration with data collection tools}

\sotas can integrate tools like MTData, which automates the collection and preparation of machine translation data sets \cite{gowda-etal-2021-many}.
The following example shows \texttt{mtdata} pipeline which downloads the specified data sets and mixes them as per \texttt{$--$mix-weights} argument:

\begin{lstlisting}[language=bash,basicstyle=\ttfamily\small]
sotastream -n 1 mtdata --langs rus-eng \
  Statmt-news_commentary-16-eng-rus \
  Statmt-backtrans_ruen-wmt20-rus-eng \
  OPUS-paracrawl-v9-eng-rus \
  --mix-weights 2 1 1
\end{lstlisting}

\subsection{Generating data sets for offline use}

If the training tool does not support consuming training data from the standard input, \sotas can be used for static data generation. 
While the real advantages of \sotas accrue when making use of its on-the-fly data manipulations, this approach retains some of its benefits.

\subsection{Other uses}

The \sotas approach to factoring data generation, as well as \sotas itself, could also be used for generating non-textual content.
The benefits of not writing data to disk would be greater in settings where input disk space is larger than plain text, such as translation from visual representations \cite{salesky-etal-2021-robust}.
Nor does it need to be limited to sequence-to-sequence settings; we imagine the approach could be useful for training of LLMs.

\section{Related Work}

To our knowledge, \sotas is novel in presenting a framework for the generation of training data as a distinct component in the model training pipeline.
It emphasizes a clean separation between data generation and training, multithreading for throughput, and the use of the standard \unix pipeline interface.

There are many libraries focused on data augmentation.
A number of these are focused just on text augmentations, including nlpaug \cite{ma-2019-nlpaug}, TextAttack \cite{morris-etal-2020-textattack}, and TextFlint \cite{gui-etal-2021-textflint}.
Another tool is AugLy \cite{papakipos-bitton-2022-augly}, a multimodal tool for text, audio, images, and video that provides robust training against adversarial perturbations.
Many of these libraries would be useful within \sotas's general framework.

\section{Conclusion}
\label{section:conclusion}

Working with large datasets can be difficult.
It is time-consuming to copy and process data, and can be expensive to store to disk.
The data-preprocessing approach that is common in machine translation model training makes it possible to work with increasingly large datasets, but this ability does not come without costs.
If data is compiled with model-specific parameters that tie the data to a particular model training, it prevents or at least complicates reusability.
This problem is further exacerbated by research settings where one of the experimental parameters is manipulations of the training data, since each variant must be written to disk and then managed.

We have described an approach that separates data generation from data consumption, and shared \sotas, an implementation that makes use of the standard \unix pipeline.
The requirement is that preprocessing must now be computed on the fly.
Our experiments show that this does not slow down training, nor does it affect the accuracy of the models trained.
The approach provides flexibility, saves processing time and disk space, and simplifies experiment management.

\section*{Limitations}

We have only investigated data consumption rates in a single toolkit, Marian, written in C++.
It's possible that the online preprocessing requirements may be too much for toolkits written in languages without a proper thread implementation.

\balance

\bibliography{anthology,custom}

\begin{thebibliography}{25}
\expandafter\ifx\csname natexlab\endcsname\relax\def\natexlab#1{#1}\fi

\bibitem[{Ba{\~n}{\'o}n et~al.(2020)Ba{\~n}{\'o}n, Chen, Haddow, Heafield,
  Hoang, Espl{\`a}-Gomis, Forcada, Kamran, Kirefu, Koehn, Ortiz~Rojas,
  Pla~Sempere, Ram{\'\i}rez-S{\'a}nchez, Sarr{\'\i}as, Strelec, Thompson,
  Waites, Wiggins, and Zaragoza}]{banon-etal-2020-paracrawl}
Marta Ba{\~n}{\'o}n, Pinzhen Chen, Barry Haddow, Kenneth Heafield, Hieu Hoang,
  Miquel Espl{\`a}-Gomis, Mikel~L. Forcada, Amir Kamran, Faheem Kirefu, Philipp
  Koehn, Sergio Ortiz~Rojas, Leopoldo Pla~Sempere, Gema
  Ram{\'\i}rez-S{\'a}nchez, Elsa Sarr{\'\i}as, Marek Strelec, Brian Thompson,
  William Waites, Dion Wiggins, and Jaume Zaragoza. 2020.
\newblock \href {https://doi.org/10.18653/v1/2020.acl-main.417} {{P}ara{C}rawl:
  Web-scale acquisition of parallel corpora}.
\newblock In \emph{Proceedings of the 58th Annual Meeting of the Association
  for Computational Linguistics}, pages 4555--4567, Online. Association for
  Computational Linguistics.

\bibitem[{Bergmanis and Pinnis(2021)}]{bergmanis-pinnis-2021-facilitating}
Toms Bergmanis and M{\=a}rcis Pinnis. 2021.
\newblock \href {https://doi.org/10.18653/v1/2021.eacl-main.271} {Facilitating
  terminology translation with target lemma annotations}.
\newblock In \emph{Proceedings of the 16th Conference of the European Chapter
  of the Association for Computational Linguistics: Main Volume}, pages
  3105--3111, Online. Association for Computational Linguistics.

\bibitem[{Chen et~al.(2016)Chen, Matusov, Khadivi, and
  Peter}]{chen-etal-2016-guided}
Wenhu Chen, Evgeny Matusov, Shahram Khadivi, and Jan-Thorsten Peter. 2016.
\newblock \href {https://aclanthology.org/2016.amta-researchers.10} {Guided
  alignment training for topic-aware neural machine translation}.
\newblock In \emph{Conferences of the Association for Machine Translation in
  the Americas: MT Researchers' Track}, pages 121--134, Austin, TX, USA. The
  Association for Machine Translation in the Americas.

\bibitem[{Gowda et~al.(2021)Gowda, Zhang, Mattmann, and
  May}]{gowda-etal-2021-many}
Thamme Gowda, Zhao Zhang, Chris Mattmann, and Jonathan May. 2021.
\newblock \href {https://doi.org/10.18653/v1/2021.acl-demo.37}
  {Many-to-{E}nglish machine translation tools, data, and pretrained models}.
\newblock In \emph{Proceedings of the 59th Annual Meeting of the Association
  for Computational Linguistics and the 11th International Joint Conference on
  Natural Language Processing: System Demonstrations}, pages 306--316, Online.
  Association for Computational Linguistics.

\bibitem[{Gui et~al.(2021)Gui, Wang, Zhang, Liu, Zou, Zhou, Zheng, Zhang, Wu,
  Ye, Pang, Zhang, Li, Ma, Fei, Cai, Zhao, Hu, Yan, Tan, Hu, Bian, Liu, Zhu,
  Qin, Xing, Fu, Zhang, Peng, Zheng, Zhou, Wei, Qiu, and
  Huang}]{gui-etal-2021-textflint}
Tao Gui, Xiao Wang, Qi~Zhang, Qin Liu, Yicheng Zou, Xin Zhou, Rui Zheng, Chong
  Zhang, Qinzhuo Wu, Jiacheng Ye, Zexiong Pang, Yongxin Zhang, Zhengyan Li,
  Ruotian Ma, Zichu Fei, Ruijian Cai, Jun Zhao, Xinwu Hu, Zhiheng Yan, Yiding
  Tan, Yuan Hu, Qiyuan Bian, Zhihua Liu, Bolin Zhu, Shan Qin, Xiaoyu Xing,
  Jinlan Fu, Yue Zhang, Minlong Peng, Xiaoqing Zheng, Yaqian Zhou, Zhongyu Wei,
  Xipeng Qiu, and Xuanjing Huang. 2021.
\newblock \href {http://arxiv.org/abs/2103.11441} {Textflint: Unified
  multilingual robustness evaluation toolkit for natural language processing}.
\newblock \emph{CoRR}, abs/2103.11441.

\bibitem[{Hieber et~al.(2022)Hieber, Denkowski, Domhan, Barros, Ye, Niu, Hoang,
  Tran, Hsu, Nadejde, Lakew, Mathur, Currey, and
  Federico}]{hieber-etal-2022-sockeye}
Felix Hieber, Michael Denkowski, Tobias Domhan, Barbara~Darques Barros,
  Celina~Dong Ye, Xing Niu, Cuong Hoang, Ke~Tran, Benjamin Hsu, Maria Nadejde,
  Surafel Lakew, Prashant Mathur, Anna Currey, and Marcello Federico. 2022.
\newblock \href {http://arxiv.org/abs/2207.05851} {Sockeye 3: Fast neural
  machine translation with pytorch}.

\bibitem[{Jalili~Sabet et~al.(2020)Jalili~Sabet, Dufter, Yvon, and
  Sch{\"u}tze}]{jalili-sabet-etal-2020-simalign}
Masoud Jalili~Sabet, Philipp Dufter, Fran{\c{c}}ois Yvon, and Hinrich
  Sch{\"u}tze. 2020.
\newblock \href {https://doi.org/10.18653/v1/2020.findings-emnlp.147}
  {{S}im{A}lign: High quality word alignments without parallel training data
  using static and contextualized embeddings}.
\newblock In \emph{Findings of the Association for Computational Linguistics:
  EMNLP 2020}, pages 1627--1643, Online. Association for Computational
  Linguistics.

\bibitem[{Junczys-Dowmunt et~al.(2018)Junczys-Dowmunt, Heafield, Hoang,
  Grundkiewicz, and Aue}]{junczys-dowmunt-etal-2018-marian-cost}
Marcin Junczys-Dowmunt, Kenneth Heafield, Hieu Hoang, Roman Grundkiewicz, and
  Anthony Aue. 2018.
\newblock \href {https://doi.org/10.18653/v1/W18-2716} {{M}arian:
  Cost-effective high-quality neural machine translation in {C}++}.
\newblock In \emph{Proceedings of the 2nd Workshop on Neural Machine
  Translation and Generation}, pages 129--135, Melbourne, Australia.
  Association for Computational Linguistics.

\bibitem[{Kocmi et~al.(2022)Kocmi, Bawden, Bojar, Dvorkovich, Federmann,
  Fishel, Gowda, Graham, Grundkiewicz, Haddow, Knowles, Koehn, Monz, Morishita,
  Nagata, Nakazawa, Nov{\'a}k, Popel, and
  Popovi{\'c}}]{kocmi-etal-2022-findings}
Tom Kocmi, Rachel Bawden, Ond{\v{r}}ej Bojar, Anton Dvorkovich, Christian
  Federmann, Mark Fishel, Thamme Gowda, Yvette Graham, Roman Grundkiewicz,
  Barry Haddow, Rebecca Knowles, Philipp Koehn, Christof Monz, Makoto
  Morishita, Masaaki Nagata, Toshiaki Nakazawa, Michal Nov{\'a}k, Martin Popel,
  and Maja Popovi{\'c}. 2022.
\newblock \href {https://aclanthology.org/2022.wmt-1.1} {Findings of the 2022
  conference on machine translation ({WMT}22)}.
\newblock In \emph{Proceedings of the Seventh Conference on Machine Translation
  (WMT)}, pages 1--45, Abu Dhabi, United Arab Emirates (Hybrid). Association
  for Computational Linguistics.

\bibitem[{Kudo(2018)}]{kudo-2018-subword}
Taku Kudo. 2018.
\newblock \href {https://doi.org/10.18653/v1/P18-1007} {Subword regularization:
  Improving neural network translation models with multiple subword
  candidates}.
\newblock In \emph{Proceedings of the 56th Annual Meeting of the Association
  for Computational Linguistics (Volume 1: Long Papers)}, pages 66--75,
  Melbourne, Australia. Association for Computational Linguistics.

\bibitem[{Kudo and Richardson(2018)}]{kudo-richardson-2018-sentencepiece}
Taku Kudo and John Richardson. 2018.
\newblock \href {https://doi.org/10.18653/v1/D18-2012} {{S}entence{P}iece: A
  simple and language independent subword tokenizer and detokenizer for neural
  text processing}.
\newblock In \emph{Proceedings of the 2018 Conference on Empirical Methods in
  Natural Language Processing: System Demonstrations}, pages 66--71, Brussels,
  Belgium. Association for Computational Linguistics.

\bibitem[{Li et~al.(2019)Li, Michel, Anastasopoulos, Belinkov, Durrani, Firat,
  Koehn, Neubig, Pino, and Sajjad}]{li-etal-2019-findings}
Xian Li, Paul Michel, Antonios Anastasopoulos, Yonatan Belinkov, Nadir Durrani,
  Orhan Firat, Philipp Koehn, Graham Neubig, Juan Pino, and Hassan Sajjad.
  2019.
\newblock \href {https://doi.org/10.18653/v1/W19-5303} {Findings of the first
  shared task on machine translation robustness}.
\newblock In \emph{Proceedings of the Fourth Conference on Machine Translation
  (Volume 2: Shared Task Papers, Day 1)}, pages 91--102, Florence, Italy.
  Association for Computational Linguistics.

\bibitem[{Ma(2019)}]{ma-2019-nlpaug}
Edward Ma. 2019.
\newblock Nlp augmentation.
\newblock https://github.com/makcedward/nlpaug.

\bibitem[{Morris et~al.(2020)Morris, Lifland, Yoo, Grigsby, Jin, and
  Qi}]{morris-etal-2020-textattack}
John Morris, Eli Lifland, Jin~Yong Yoo, Jake Grigsby, Di~Jin, and Yanjun Qi.
  2020.
\newblock \href {https://doi.org/10.18653/v1/2020.emnlp-demos.16}
  {{T}ext{A}ttack: A framework for adversarial attacks, data augmentation, and
  adversarial training in {NLP}}.
\newblock In \emph{Proceedings of the 2020 Conference on Empirical Methods in
  Natural Language Processing: System Demonstrations}, pages 119--126, Online.
  Association for Computational Linguistics.

\bibitem[{Nowakowski et~al.(2022)Nowakowski, Pa{\l}ka, Guttmann, and
  Pokrywka}]{nowakowski-etal-2022-adam}
Artur Nowakowski, Gabriela Pa{\l}ka, Kamil Guttmann, and Miko{\l}aj Pokrywka.
  2022.
\newblock \href {https://aclanthology.org/2022.wmt-1.26} {{A}dam {M}ickiewicz
  {U}niversity at {WMT} 2022: {NER}-assisted and quality-aware neural machine
  translation}.
\newblock In \emph{Proceedings of the Seventh Conference on Machine Translation
  (WMT)}, pages 326--334, Abu Dhabi, United Arab Emirates (Hybrid). Association
  for Computational Linguistics.

\bibitem[{Ott et~al.(2019)Ott, Edunov, Baevski, Fan, Gross, Ng, Grangier, and
  Auli}]{ott-etal-2019-fairseq}
Myle Ott, Sergey Edunov, Alexei Baevski, Angela Fan, Sam Gross, Nathan Ng,
  David Grangier, and Michael Auli. 2019.
\newblock \href {https://doi.org/10.18653/v1/N19-4009} {fairseq: A fast,
  extensible toolkit for sequence modeling}.
\newblock In \emph{Proceedings of the 2019 Conference of the North {A}merican
  Chapter of the Association for Computational Linguistics (Demonstrations)},
  pages 48--53, Minneapolis, Minnesota. Association for Computational
  Linguistics.

\bibitem[{Papakipos and Bitton(2022)}]{papakipos-bitton-2022-augly}
Zoe Papakipos and Joanna Bitton. 2022.
\newblock \href {http://arxiv.org/abs/2201.06494} {Augly: Data augmentations
  for robustness}.
\newblock \emph{CoRR}, abs/2201.06494.

\bibitem[{Papineni et~al.(2002)Papineni, Roukos, Ward, and
  Zhu}]{papineni-etal-2002-bleu}
Kishore Papineni, Salim Roukos, Todd Ward, and Wei-Jing Zhu. 2002.
\newblock \href {https://doi.org/10.3115/1073083.1073135} {{B}leu: a method for
  automatic evaluation of machine translation}.
\newblock In \emph{Proceedings of the 40th Annual Meeting of the Association
  for Computational Linguistics}, pages 311--318, Philadelphia, Pennsylvania,
  USA. Association for Computational Linguistics.

\bibitem[{Popovi{\'c}(2015)}]{popovic-2015-chrf}
Maja Popovi{\'c}. 2015.
\newblock \href {https://doi.org/10.18653/v1/W15-3049} {chr{F}: character
  n-gram {F}-score for automatic {MT} evaluation}.
\newblock In \emph{Proceedings of the Tenth Workshop on Statistical Machine
  Translation}, pages 392--395, Lisbon, Portugal. Association for Computational
  Linguistics.

\bibitem[{Post(2018)}]{post-2018-call}
Matt Post. 2018.
\newblock \href {https://doi.org/10.18653/v1/W18-6319} {A call for clarity in
  reporting {BLEU} scores}.
\newblock In \emph{Proceedings of the Third Conference on Machine Translation:
  Research Papers}, pages 186--191, Brussels, Belgium. Association for
  Computational Linguistics.

\bibitem[{Post and Junczys-Dowmunt(2023)}]{post-junczys-dowmunt-2023-escaping}
Matt Post and Marcin Junczys-Dowmunt. 2023.
\newblock \href {http://arxiv.org/abs/2304.12959} {Escaping the sentence-level
  paradigm in machine translation}.

\bibitem[{Rei et~al.(2020)Rei, Stewart, Farinha, and
  Lavie}]{rei-etal-2020-comet}
Ricardo Rei, Craig Stewart, Ana~C Farinha, and Alon Lavie. 2020.
\newblock \href {https://doi.org/10.18653/v1/2020.emnlp-main.213} {{COMET}: A
  neural framework for {MT} evaluation}.
\newblock In \emph{Proceedings of the 2020 Conference on Empirical Methods in
  Natural Language Processing (EMNLP)}, pages 2685--2702, Online. Association
  for Computational Linguistics.

\bibitem[{Salesky et~al.(2021)Salesky, Etter, and
  Post}]{salesky-etal-2021-robust}
Elizabeth Salesky, David Etter, and Matt Post. 2021.
\newblock \href {https://doi.org/10.18653/v1/2021.emnlp-main.576} {Robust
  open-vocabulary translation from visual text representations}.
\newblock In \emph{Proceedings of the 2021 Conference on Empirical Methods in
  Natural Language Processing}, pages 7235--7252, Online and Punta Cana,
  Dominican Republic. Association for Computational Linguistics.

\bibitem[{Vaswani et~al.(2017)Vaswani, Shazeer, Parmar, Uszkoreit, Jones,
  Gomez, Kaiser, and Polosukhin}]{vaswani-etal-2017-attention}
Ashish Vaswani, Noam Shazeer, Niki Parmar, Jakob Uszkoreit, Llion Jones,
  Aidan~N. Gomez, Lukasz Kaiser, and Illia Polosukhin. 2017.
\newblock \href {http://arxiv.org/abs/1706.03762} {Attention is all you need}.
\newblock \emph{CoRR}, abs/1706.03762.

\bibitem[{Zeng et~al.(2021)Zeng, Liu, Li, Ran, Meng, Li, Xu, and
  Zhou}]{zeng-etal-2021-wechat}
Xianfeng Zeng, Yijin Liu, Ernan Li, Qiu Ran, Fandong Meng, Peng Li, Jinan Xu,
  and Jie Zhou. 2021.
\newblock \href {https://aclanthology.org/2021.wmt-1.23} {{W}e{C}hat neural
  machine translation systems for {WMT}21}.
\newblock In \emph{Proceedings of the Sixth Conference on Machine Translation},
  pages 243--254, Online. Association for Computational Linguistics.

\end{thebibliography}
\bibliographystyle{acl_natbib}

\end{document}